\documentclass[10pt,twocolumn,letterpaper]{article}

\usepackage[pagenumbers]{cvpr} 

\usepackage[dvipsnames]{xcolor}

\usepackage{times}
\usepackage{epsfig}
\usepackage{graphicx}
\usepackage{amsmath}
\usepackage{amssymb}
\usepackage{comment}
\usepackage{bm}
\usepackage{xcolor}

\usepackage{comment}
\usepackage{multirow}
\usepackage[subrefformat=parens]{subcaption}
\usepackage{colortbl}

\usepackage[pagebackref,breaklinks,colorlinks]{hyperref}

\def\paperID{281} 
\def\confName{3DV\xspace}
\def\confYear{2024\xspace}

\title{DeepShaRM: Multi-View Shape and Reflectance Map Recovery\\Under Unknown Lighting}

\author{Kohei Yamashita \qquad\qquad Shohei Nobuhara \qquad\qquad Ko Nishino\\
Graduate School of Informatics, Kyoto University,
Kyoto, Japan\\
{\tt\small \url{https://vision.ist.i.kyoto-u.ac.jp/}}
}

\begin{document}

\twocolumn[{
\maketitle
\begin{center}
    \captionsetup{type=figure}
    \includegraphics[width=\linewidth]{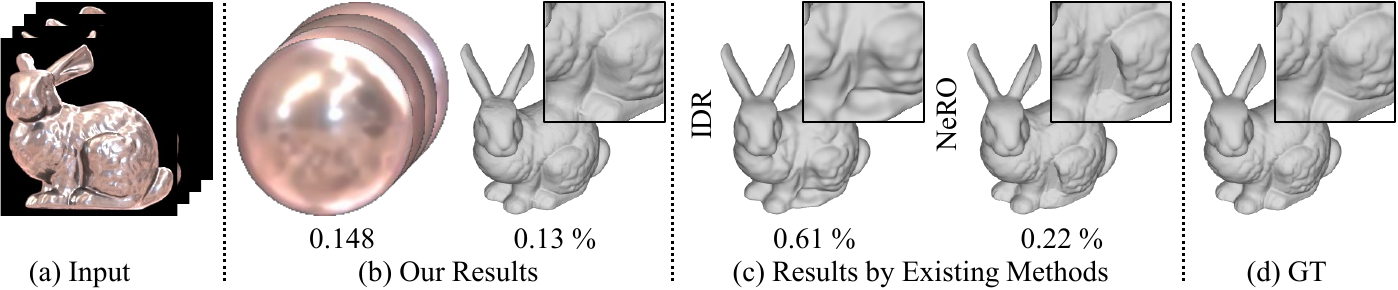}
        \captionof{figure}{We introduce a novel 3D reconstruction method that jointly estimates camera-view reflectance maps and surface geometry (b) of textureless, non-Lambertian objects from posed multi-view images captured under unknown natural illumination (a). By directly recovering reflectance maps with learned priors instead of solving the ill-posed decomposition of surface appearance into reflectance and illumination, our method achieves state-of-the-art geometry reconstruction accuracy even for objects with non-Lambertian reflectance and under complex natural illumination, which remains challenging for existing methods (c). The ground truth geometry is shown in (d). Each inset figure shows the detail around the leg. The log-scale mean absolute error for the reflectance maps and the root-mean-square errors for the 3D shapes are also shown.}
    \label{fig:opening}
\end{center}
}]
\thispagestyle{empty}

\begin{abstract}
    Geometry reconstruction of textureless, non-Lambertian objects under unknown natural illumination (\ie, in the wild) remains challenging as correspondences cannot be established and the reflectance cannot be expressed in simple analytical forms. We derive a novel multi-view method, DeepShaRM, that achieves state-of-the-art accuracy on this challenging task. Unlike past methods that formulate this as inverse-rendering, \ie, estimation of reflectance, illumination, and geometry from images, our key idea is to realize that reflectance and illumination need not be disentangled and instead estimated as a compound reflectance map. We introduce a novel deep reflectance map estimation network that recovers the camera-view reflectance maps from the surface normals of the current geometry estimate and the input multi-view images. The network also explicitly estimates per-pixel confidence scores to handle global light transport effects. A deep shape-from-shading network then updates the geometry estimate expressed with a signed distance function using the recovered reflectance maps. By alternating between these two, and, most important, by bypassing the ill-posed problem of reflectance and illumination decomposition, the method accurately recovers object geometry in these challenging settings. Extensive experiments on both synthetic and real-world data clearly demonstrate its state-of-the-art accuracy.  
    \end{abstract}

\section{Introduction}

Three-dimensional shape reconstruction has been a central research topic in computer vision since its inception. Diverse research efforts in the field has led to enormous progress that collectively have, more or less, solved 3D reconstruction of textured Lambertian objects. Texture provides the basis for correspondence matching and triangulation, and Lambertian reflectance lends a convenient linear expression of object appearance, both of which can be exploited in well-established approaches including stereo, multi-view stereo, and photometric stereo. Recent learning-based approaches let us also handle reasonable deviations from these ideals through the variations in the training data.

Textureless non-Lambertian surfaces, however, still remain challenging as the object appearance becomes view-dependent without visual cues for direct correspondence establishment. Several works have tackled this challenging problem, mainly by fully decomposing the object appearance into its radiometric roots, namely illumination, reflectance, and geometry. By ``inverse-rendering'' the images, the object geometry as a physical component of the rendering equation~\cite{kajiya86rendering} can be recovered. This, however, amounts to jointly solving for the illumination and reflectance. This disentanglement is highly ill-posed~\cite{ramamoorthi2004freq, lombardi2016rani, chen2020ibrdf} even when the object geometry is known as it suffers from fundamental ambiguities in frequency and color. For instance, a white sharp lighting will produce the same appearance for a red object with low-frequency reflectance characteristics as a white shiny object taken under a red blurry lighting. 

Our key observation is that we do not need to estimate reflectance and illumination as long as we can recover \textit{reflectance maps}~\cite{horn1979reflectancemap}, view-dependent mappings from a surface normal to the observed radiance that encode the compound radiometric effects of both. In fact, a recent work by Yamashita \etal~\cite{yamashita2023nlmvs} achieves geometry reconstruction of textureless, non-Lambertian objects by exploiting reflectance maps as additional inputs of a deep multi-view stereo network. This method, however, assumes known illumination and cannot be applied to images captured in the wild. 

In this paper, we introduce a novel method for recovering the geometry of textureless, non-Lambertian objects captured in multi-view images taken under unknown illumination. Our key idea, inspired by \cite{yamashita2023nlmvs}, is to completely avoid solving the ill-posed problem of appearance decomposition into reflectance and illumination but instead fully leverage camera-view reflectance maps (RMs) for geometry recovery. This is, however, nontrivial for unknown illumination as, although the ambiguity between reflectance and illumination is bypassed, the ambiguity between geometry and reflectance maps still remains. 

We resolve this with a canonical joint estimation framework by introducing two deep neural networks that each leverage strong priors on geometry and reflectance maps learnable from large-scale synthetic data. From a visual hull computed from the multi-view input images, a deep reflectance map estimation network (RM network) estimates camera-view reflectance maps from input images and surface normal maps extracted from the current geometry estimate. The recovered reflectance maps and the input images are then fed into a deep shape-from-shading network (SfS network) that estimates per-pixel surface normals for each view. We update the current geometry estimate using the estimated surface normals and iterate until convergence. 

When estimating the reflectance map from the current (inaccurate) geometry estimate, pixels with global illumination effects (\ie, shadows and interreflections) and sharp changes of surface normal within its span deviate from the reflectance maps. We introduce two novel ideas to solve this problem. First, we introduce a feature extraction sub-network before the reflectance map computation so that it can implicitly learn to refine pixels inconsistent with the reflectance map. Second, when mapping the pixel to a point on the reflectance map, we explicitly filter them by estimating pixel-wise confidence scores of observations. To achieve this, we derive a differentiable weighted mapping from an image feature map to a feature reflectance map, which enables end-to-end training.

Once a reflectance map is recovered for each view, per-view, per-pixel surface normals can be recovered by a deep SfS network which leverages pixel-wise likelihoods of surface normals computed using the reflectance maps~\cite{yamashita2023nlmvs}. The estimates, however, do not represent the surface geometry itself and can be inconsistent across views. We integrate the per-view surface normal estimates by deriving a differentiable shape optimization method that updates a single geometry estimate with them. We represent the geometry estimate as a signed distance function so that objects with unknown topology can be accurately expressed. 

\begin{table}[t]
  \centering
  \setlength{\tabcolsep}{2.7pt}
  \footnotesize
  \def\HEAD#1#2{\begin{tabular}{c}#1\\#2\end{tabular}}
  \begin{tabular}{l|ccc}
     &\HEAD{ w/o }{Decomposition} & \HEAD{w/ Geometry} {Prior} & \HEAD{w/ Appearance} {Prior} \\ \hline
    IDR~\cite{yariv2020idr} & \checkmark &   \\
    NeuS~\cite{wang2021neus} & \checkmark &   \\
    PhySG~\cite{zhanf2021physg} &  &  & \checkmark  \\
    NDRMC~\cite{hasselgren2022nvdiffrecmc} &  &  & \checkmark  \\
    NeRO~\cite{liu2023nero} &  & & \checkmark  \\
    \rowcolor[rgb]{0.93,1.0,0.87} Ours & \checkmark & \checkmark & \checkmark 
  \end{tabular}
  \caption{Multi-view 3D reconstruction methods that handle non-Lambertian surfaces under unknown natural illumination. We realize accurate geometry and appearance estimation by bypassing appearance decomposition into reflectance and illumination and by leveraging learned strong priors on both geometry and appearance (reflectance maps in our method).}
  \label{tab:methods}
\end{table}

DeepShaRM enables accurate 3D shape reconstruction of objects with complex reflectance from images taken under unknown natural illumination. We extensively evaluate the effectiveness of our method on a number of experiments on synthetic and real data, and comprehensive comparison with related work. The proposed method achieves state-of-the-art accuracy. We believe our method serves as a practical means for accurate 3D shape reconstruction in the wild.

\begin{figure*}[t]
  \centering
  \includegraphics[keepaspectratio, width=\linewidth]{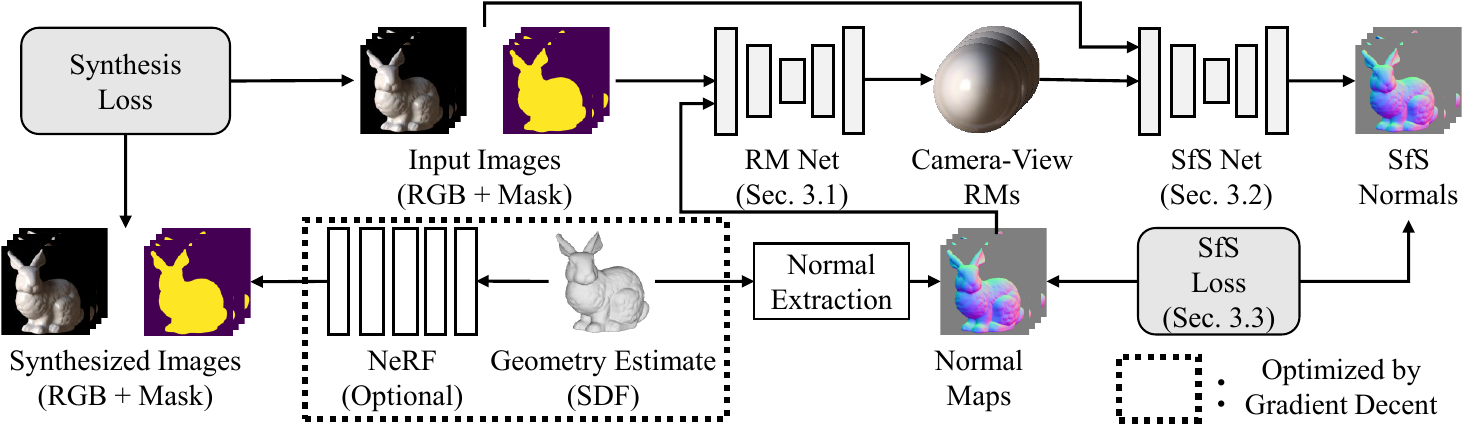}
  \caption{Overview of DeepShaRM. The inputs are posed multi-view images of a textureless, non-Lambertian object taken under unknown natural illumination and an initial coarse geometry estimate (a visual hull computed from the multi-view images). We first recover a camera-view reflectance map (RM) for each view using an input image and surface normals extracted from the current geometry estimate using a deep reflectance map estimation network (RM Net). We then feed the recovered reflectance maps along with the input images into a deep shape-from-shading network (SfS Net) that estimates per-pixel surface normals for each view. The estimated surface normals and a neural rendering method similar to NeRF~\cite{mildenhall2020nerf} (optional) are used to update the geometry estimate represented as a signed distance function (SDF). We alternate between estimating reflectance maps and updating the geometry estimate until convergence.
  }
  \label{fig:overview}
\end{figure*}

\section{Related Work}
We review relevant geometry estimation methods for textureless, non-Lambertian objects. \Cref{tab:methods} summarizes differences between our method and existing methods that also estimate the geometry of non-Lambertian objects under unknown natural illumination.

\paragraph{Inverse rendering} methods recover geometry by inverting the radiometric image formation process~\cite{oxholm2015shape, cheng2021mv3d, bi2020deep3d, zhang2022iron, nam2018svbrdf}. While traditional methods make restricting assumptions on imaging (\eg, known illumination), a few recent methods jointly estimate geometry, reflectance, and illumination from multi-view images captured in the wild~\cite{zhanf2021physg, hasselgren2022nvdiffrecmc, liu2023nero}. They leverage a microfacet BRDF model as a strong constraint on appearance and jointly estimate the physical components of the rendering equation~\cite{kajiya86rendering} in a differentiable manner. These methods, however, require dense (\eg, 100) view sampling to handle textureless, non-Lambertian surfaces as joint estimation of reflectance and illumination is extremely challenging even with known geometry~\cite{ramamoorthi2004freq, lombardi2016rani, chen2020ibrdf}. Yet they still suffer from the fundamental ambiguity between reflectance and illumination in frequency and color. We bypass these problems by estimating the compound radiometric effect of reflectance and illumination as reflectance maps. 

Georgoulis \etal~\cite{georgoulis18deeprani} introduced a learning-based reflectance map estimation method that estimates surface normals from an input image and then uses them along with the input for reflectance map recovery. This method can only handle a specific object category (car) as it relies on category-specific single-view surface normal estimation. Our method can recover accurate arbitrary objects without any category-specific prior. We also handle global illumination effects (shadows and interreflections) and sharp changes of surface normals within a single pixel.

Yamashita \etal~\cite{yamashita2023nlmvs} introduced a deep multi-view stereo network for textureless, non-Lambertian object captured under known lighting by leveraging reflectance maps as auxiliary inputs. Our method fundamentally differs in that we do not assume known illumination, which is essential for ``in-the-wild'' geometry reconstruction. Their method also only guarantees geometry consistency among neighboring views and the final geometry estimate can still be inconsistent across views. In contrast, we represent the object geometry with a single signed distance function and estimate it using per-view estimates of the SfS network while alternating with reflectance map estimation. This enables a view-independent geometry reconstruction which is consistent with all input views.

\begin{figure*}[t]
  \centering
  \includegraphics[keepaspectratio, width=0.98\linewidth]{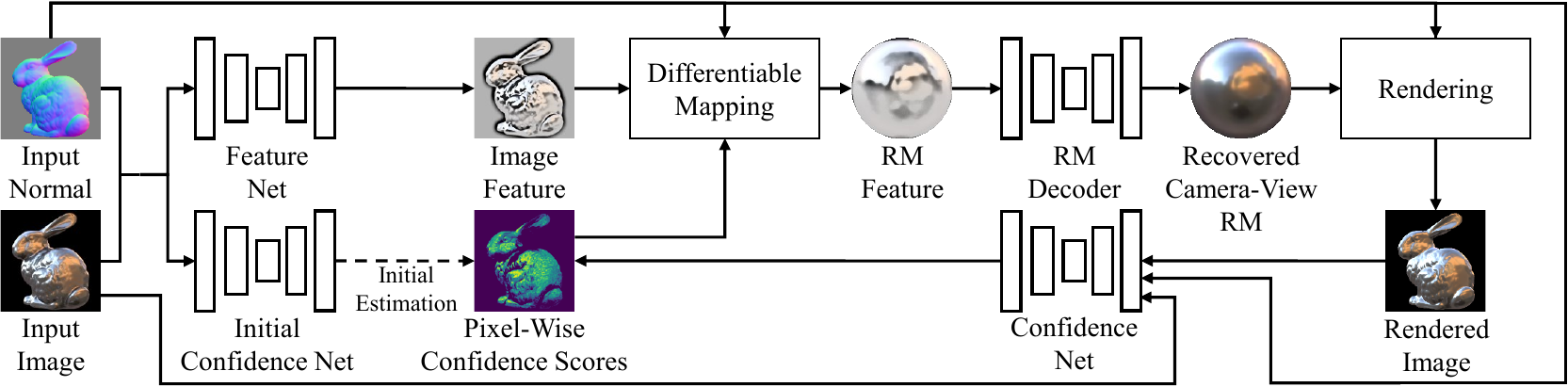}
  \caption{In RM network, we first apply an image-space feature extraction (Feature Net) to the inputs so that it can implicitly learn to refine pixels inconsistent with the camera-view reflectance map due to global illumination effects (shadows and intrreflections) and sharp changes of surface normals. We then map the extracted image feature into an feature reflectance map using the input surface normals and decode the feature reflectance map using another 2D UNet (RM Decoder). We also explicitly estimate pixel wise confidence scores and alternate with the reflectance map estimation so that the network can learn to filter pixels that suffer from the deviations. In the alternating estimation, we render an image from the recovered reflectance map and the input surface normal map for the confidence estimation.}
   \label{fig:rm_net}
\end{figure*}

\paragraph{Neural image synthesis} methods handle non-Lambertian objects by representing the surface appearance as a neural radiance field (NeRF)~\cite{mildenhall2020nerf} instead of surface reflectance and illumination. They can recover 3D geometry by jointly optimizing a NeRF and a geometry estimate using differentiable rendering~\cite{yariv2020idr, oechsle2021unisurf, yarov2021volsdf, wang2021neus}. They, however, fundamentally rely on dense (\eg, 100) view sampling as they suffer from the ambiguity between geometry and the unconstrained appearance representation. A few methods handle sparse views by conditioning the geometry and radiance field estimates with input images~\cite{long2022sparseneus, ren2023volrecon}. Directly inferring a radiance field from input images is extremely challenging and these methods struggle to deal with objects with complex appearance, \ie, textureless objects with non-Lambertian reflectance and under natural illumination.

\section{Deep Shape and Reflectance Map Recovery}

\Cref{fig:overview} shows an overview of our method which we refer to as DeepShaRM. The inputs to DeepShaRM are posed multi-view images of a textureless, non-Lambertian object captured under unknown natural illumination. We also assume that an initial coarse geometry estimate can be obtained from silhouettes of the target object. We represent the geometry estimate as a 3D grid of a signed distance function (SDF). Our goal is to recover the surface geometry of the object whose surface reflectance and surrounding illumination environment are also unknowns.

A straightforward approach to this task is to jointly estimate the surface geometry, the surface reflectance (\ie, a bidirectional reflectance distribution function, BRDF) and the illumination environment by inverting the radiometric image formation process~\cite{kajiya86rendering}. This approach, however, suffers from the fundamental ambiguity between reflectance and illumination in frequency and color~\cite{lombardi2016rani, ramamoorthi2004freq}. Also, while low-dimensional analytical models (\eg, spherical harmonics and spherical Gaussians) are often used in such inverse rendering approaches, it cannot handle complex natural illumination with high-frequency components (\eg, directional lights). We avoid these difficulties altogether by representing the compound radiometric effects of reflectance and illumination with camera-view reflectance maps~\cite{horn1979reflectancemap}. The reflectance map $R(\mathbf{n})$ is a mapping from a surface normal $\mathbf{n}$ to the observed radiance determined by the BRDF $f(\mathbf{\omega_i}, \mathbf{\omega_o}, \mathbf{n})$ and the surrounding illumination environment $L_i(\mathbf{\omega_i})$ 
\begin{equation}
    R(\mathbf{n}) = \int L_i(\mathbf{\omega_i}) f(\mathbf{\omega_i}, \mathbf{\omega_o}, \mathbf{n}) \max\left(\mathbf{\omega_i} \cdot \mathbf{n}, 0\right) \mathrm{d}\mathbf{\omega_i}\,,
\end{equation}
where $\mathbf{\omega_i}$ and $\mathbf{\omega_o}$ are incident and viewing directions. Note that we assume that the viewing direction $\mathbf{\omega_o}$ is constant for each view. 

Given the inputs, we first recover camera-view reflectance maps from surface normals of the current geometry estimate and the observed pixel values using a deep reflectance map estimation network (RM network). As the reflectance maps explicitly represent the relationship between surface appearance and surface geometry, they become tractable cues for geometry reconstruction. We use them as inputs to a deep shape-from-shading network (SfS network) that estimates pixel-wise surface normals for each view. We use the estimated surface normals to update the geometry estimate (an SDF grid) and alternate with estimation of reflectance maps. In addition to the optimization with the estimated surface normals, inspired by neural image synthesis methods~\cite{wang2021neus,yariv2020idr}, we also jointly optimize a neural radiance field (NeRF)~\cite{mildenhall2020nerf} and the SDF estimate so that errors between the images synthesized from them and the inputs are minimized. Although the NeRF is not essential for our key idea, we empirically found that it improves reconstruction of surface details (please also see ablation study in \cref{sec:exp-joint-est}). By iterating the alternating estimation until convergence, we can recover accurate geometry and reflectance maps consistent with each other.

\subsection{RM Network}

\begin{figure*}[t]
  \centering
  \includegraphics[keepaspectratio, width=\linewidth]{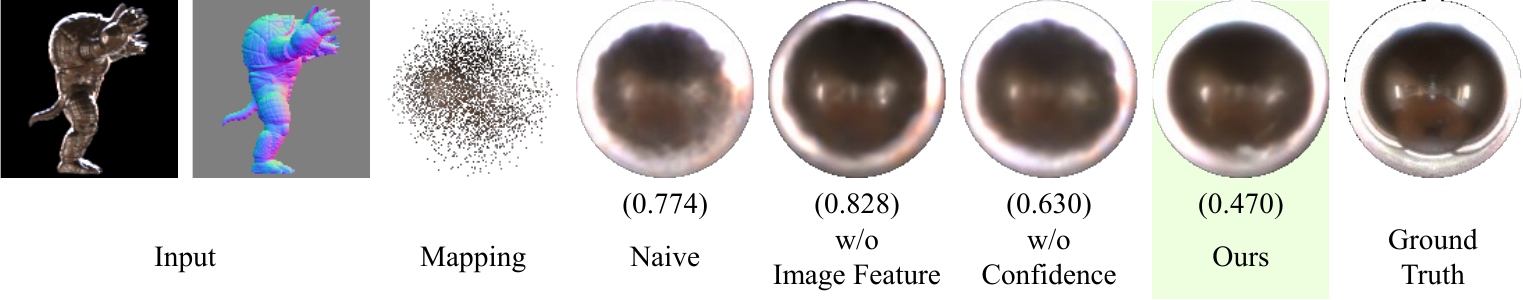}
  \caption{Example estimation results of RM Net. The numbers are log-scale mean absolute errors (lower is better). The results show the effectiveness of the proposed architecture. Please see the text for details of each method.
  }
  \label{fig:rm-results}
\end{figure*}

The first step of DeepShaRM is to estimate a reflectance map for each view given an input image and the surface normals of the current geometry estimate. Even with the geometry estimate, reflectance map recovery is challenging as the observations are sparse. As shown in the third column of \cref{fig:rm-results}, even if we map pixel values in the input image using the ground truth surface normals, dense reflectance maps cannot be recovered. Also, reconstruction errors of the current geometry estimate strongly affect the accuracy of such a naive mapping method. For these, we estimate each of these camera-view reflectance maps with a deep neural network (RM network) which learns strong priors on reflectance maps from a large scale synthetic dataset.

\Cref{fig:rm_net} depicts the architecture of the RM network. Inspired by Georgoulis \etal~\cite{georgoulis18deeprani}, we explicitly map pixels in the input image onto the spherical domain using the input surface normals and then interpolate (refine) them with a 2D convolutional neural network. A naive mapping, however, fails to handle the complex object geometry. Pixel values are often inconsistent with the reflectance map due to global illumination effects (shadows and interreflections) and sharp changes of surface normals within a single pixel. We explicitly handle such deviations with two novel ideas. 

First, we introduce an image feature extraction sub-network (Feature Net) before the mapping. By leveraging contextual information, the network can implicitly learn to refine pixels inconsistent with the reflectance map and extract additional features for the reflectance map recovery. Second, we introduce confidence score estimation sub-networks that can learn to filter pixels that suffer from the deviations. At first, we estimate pixel-wise confidence scores (weights) of observations from the input image and the input normal map using a deep neural sub-network (Initial Confidence Net). We then use them to weight the image features when mapping. Specifically, we derive a differentiable weighted mapping from an image feature map $I'_\mathbf{m}$ to a reflectance map feature $R'(\mathbf{n'})$ with pixel-wise surface normals $\mathbf{n}_\mathbf{m}$ and estimated scores $w_\mathbf{m}$
\begin{equation}
    R'(\mathbf{n'}) = \frac{\sum_{\mathbf{m}} w'_\mathbf{m} I'_\mathbf{m}}{\sum_{\mathbf{m}} w'_\mathbf{m}} \,,
    \label{eq:mapping}
\end{equation}
where 
\begin{equation}
    w'_\mathbf{m} \equiv w_\mathbf{m} \max\left(\mathbf{n}_\mathbf{m} \cdot \mathbf{\omega_o}, 0\right) \exp\left(s \mathbf{n_m} \cdot \mathbf{n'}\right) \,.
\end{equation}

In addition to $w_\mathbf{m}$, we use weights determined by distances between the input normal $\mathbf{n_m}$, the query normal $\mathbf{n'}$, and the viewing direction $\mathbf{\omega_o}$ for interpolation of the sparse inputs. $s$ is a parameter of the interpolation optimized during training. In practice, we use the angular fisheye projection to treat the reflectance maps as 2D images. We compute \cref{eq:mapping} for each pixel in the projected reflectance map. A reflectance map estimate is decoded from the RM feature using another 2D convolutional sub-network (RM decoder). 

Once the camera-view reflectance map is recovered, we can detect deviations of pixels from the reflectance map by simply comparing the input image and one rendered from the reflectance map and the input normal map. We leverage this by alternating between refining the pixel-wise scores and estimating a reflectance map. In the alternating estimation, we use another score estimation sub-network (Confidence Net) that takes in the rendered image in addition to the original inputs.

We train the proposed RM network in a supervised manner. We use a synthetic dataset for training, so that ground truth reflectance maps are available. We train the network by minimizing errors between estimated reflectance maps and the ground truth ones. As the overall network including the mapping is differentiable, we can train the RM network in an end-to-end manner. Please see the supplemental for more details of the loss function.

\begin{figure}[t]
  \centering
  \includegraphics[keepaspectratio, width=\linewidth]{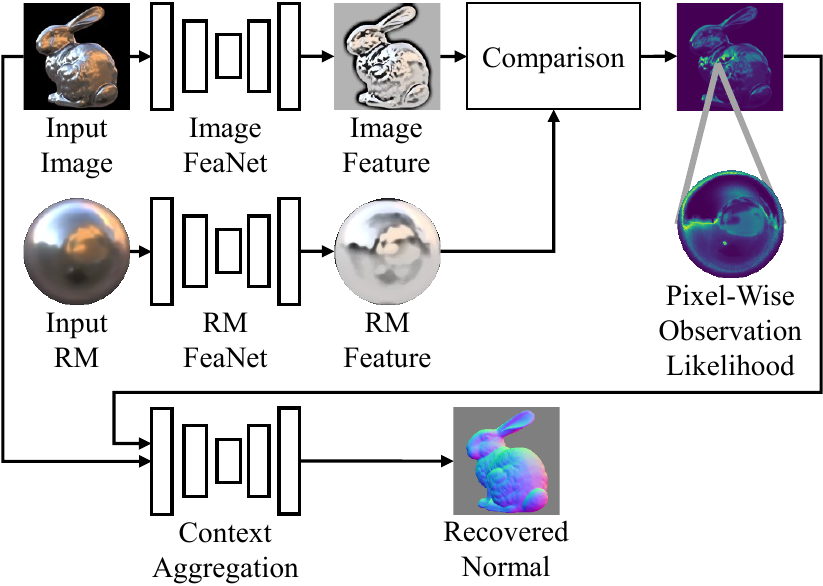}
  \caption{Overview of the architecture of the shape-from-shading network. Inspired by Yamashita \etal~\cite{yamashita2023nlmvs}, we compute a pixel-wise observation likelihood for each surface normal candidate and, by using the input image and the input reflectance map, improve consistency between the inputs and the output. 
  }
  \label{fig:sfs-net}
\end{figure}

\subsection{SfS Network}

\begin{figure*}[t]
  \centering
  \includegraphics[keepaspectratio, width=\linewidth]{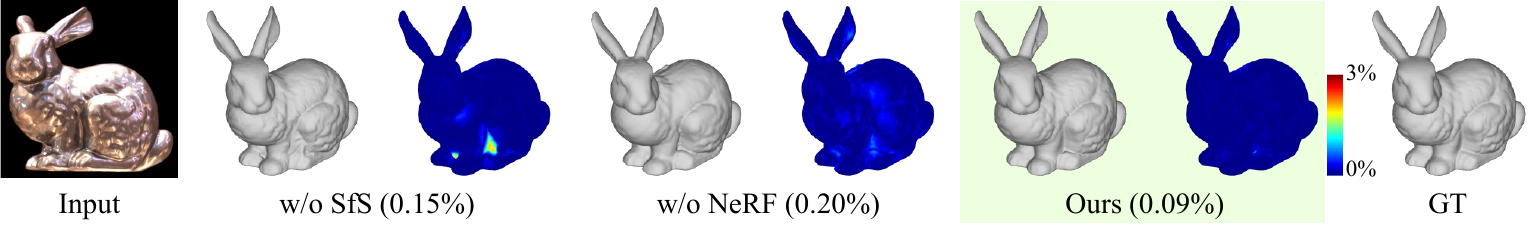}
  \caption{Ablation study of DeepShaRM on the nLMVS-Synth Dataset~\cite{yamashita2023nlmvs}. Color maps show distances to the ground truth and the numbers are the corresponding RMSE values. The results demonstrate the effectiveness of each component.
  }
  \label{fig:ablation-results}
\end{figure*}

Once camera-view reflectance maps are recovered, we can use them for estimating the surface geometry. We realize this with a deep shape-from-shading (SfS) network so that we can also leverage a learned prior on geometry. \Cref{fig:sfs-net} depicts an overview of the architecture of the SfS network. It takes in an input image and its camera-view reflectance map as inputs and estimates per-pixel surface normals for each view. Inspired by Yamashita \etal~\cite{yamashita2023nlmvs}, we compute pixel-wise observation likelihood for each surface normal candidate by comparing features extracted from the input image and the input reflectance map. By filtering the observation likelihoods and the input image using another context aggregation network, we can recover per-pixel surface normals consistent with the inputs. Similar to the RM network, we train the SfS network with a synthetic dataset in a supervised manner. 

\subsection{Joint Estimation}

Although the surface normals recovered by the SfS network capture details of the surface geometry, they are only surface normals (\ie, absolute depth cannot be recovered from a single-view estimate) which can be inconsistent across views. We integrate the per-view estimates by updating a single, view-consistent geometry estimate represented with a 3D grid of signed distance function (SDF). Inspired by Neural Implicit Evolution~\cite{mehta2022nie}, we realize this with a level set method and NVDiffRast~\cite{laine20nvdiffrast}, a differentiable renderer for 3D mesh models. We first extract a 3D mesh model from the SDF grid using Marching Cubes~\cite{lorensen87marchingcubes} and then render surface normal maps using NVDiffRast. We use the cubic B-spline interpolation so that the SDF values used for the Marching Cubes become smooth~\cite{vicini2022sdf}. We evaluate the mean absolute error between surface normals estimated by the SfS network and those rendered from the current geometry estimate as the SfS loss. We compute its gradients with respect to each vertex of the extracted mesh model using automatic differentiation of NVDiffRast and then obtain gradients with respect to parameters of the SDF using the level set equation. We update the SDF parameters using the obtained gradients and the Adam optimizer. Inspired by neural image synthesis methods~\cite{yariv2020idr, wang2021neus}, optionally, we also jointly optimize a neural radiance field~\cite{mildenhall2020nerf} along with the SDF grid so that images synthesized from them approximate the inputs. We alternate between estimating reflectance maps and updating the geometry estimate until convergence so that the geometry and reflectance map estimates become consistent with each other. Please see the supplemental for more details.

\begin{figure*}[t]
  \centering
  \includegraphics[keepaspectratio, width=\linewidth]{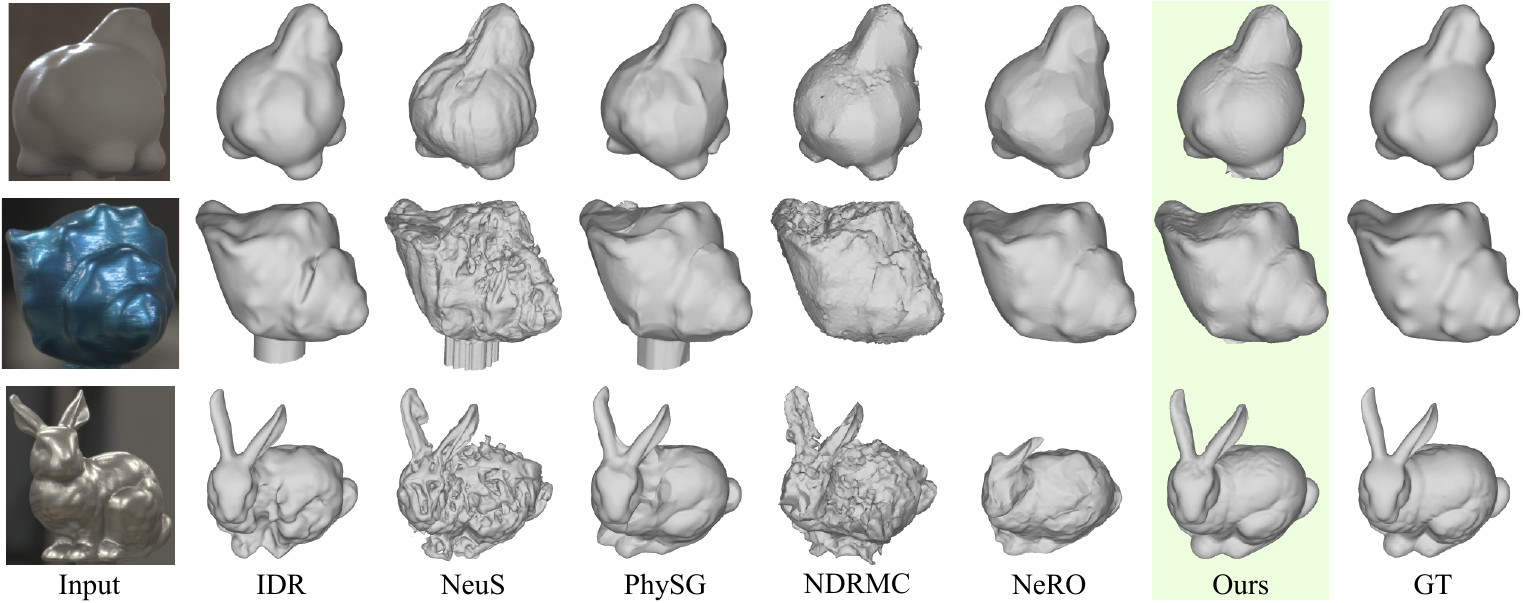}
  \caption{Geometry estimation results on nLMVS-Real Dataset~\cite{yamashita2023nlmvs}. The results are qualitatively accurate.
  }
  \label{fig:geometry-results-real}
\end{figure*}

\begin{figure}[t]
  \centering
  \includegraphics[keepaspectratio, width=\linewidth]{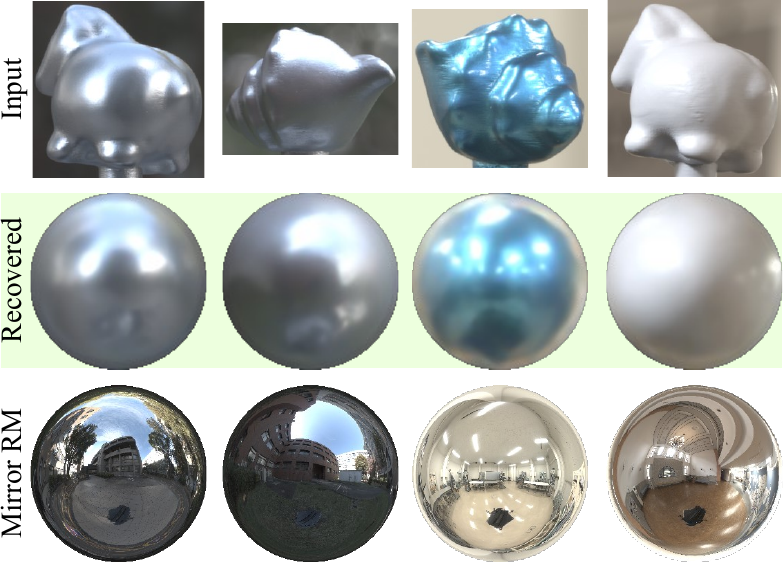}
  \caption{Recovered reflectance maps from real-world images of the nLMVS-Real dataset~\cite{yamashita2023nlmvs}. Although the ground truth reflectance maps are not available, the results are consistent with reflectance maps of mirror objects under the same environment (Mirror RM) created from ground truth illumination maps.
  }
  \label{fig:rm-results-real}
\end{figure}

\begin{figure}[t]
  \centering
  \includegraphics[keepaspectratio, width=\linewidth]{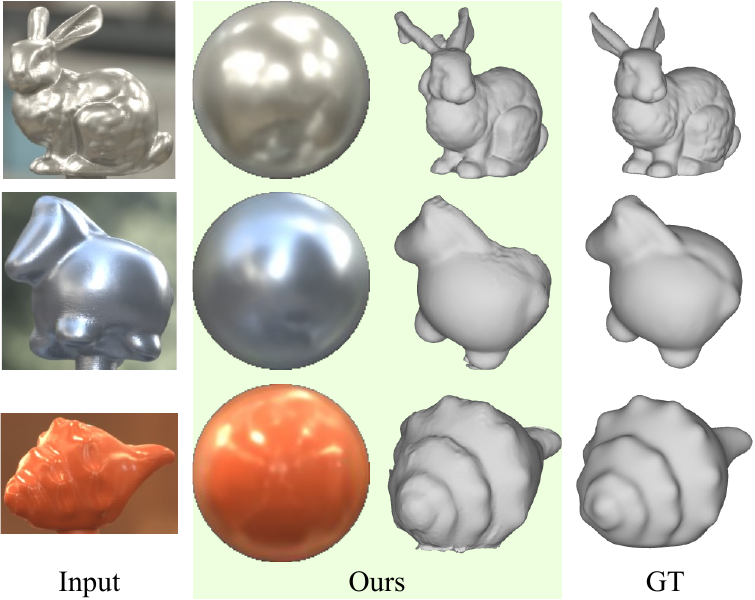}
  \caption{Recovered geometry and reflectance maps from 5-view images in the nLMVS-Real dataset~\cite{yamashita2023nlmvs}. Our method works well even with very sparse inputs.
  }
  \label{fig:geometry-results-sparse}
\end{figure}

\begin{table}[t]
  \centering
  \setlength{\tabcolsep}{2.7pt}
  \small
  \def\HEAD#1#2{\begin{tabular}{c}#1\\#2\end{tabular}}
  \begin{tabular}{l|c}
     & Log-MAE \\ \hline
    Naive & 0.162   \\
    w/o Image Feature & 0.125   \\
    w/o Confidence & 0.135 \\
    \rowcolor[rgb]{0.93,1.0,0.87} Ours & \textbf{0.121} 
  \end{tabular}
  \caption{Accuracy (Log-MAE) of recovered reflectance maps on the nLMVS-Synth dataset \cite{yamashita2023nlmvs}.}
  \label{tab:rm-results}
\end{table}

\begin{table}[t]
  \centering
  \footnotesize
  \subfloat[][]{
      \setlength{\tabcolsep}{3pt}
    \begin{tabular}{l|crr}
        & RMS1 & RMS2 \\ \hline
        IDR~\cite{yariv2020idr} & 0.26 \% & 0.27 \% \\
        NeuS~\cite{wang2021neus} & 0.96 \% & 0.66 \% \\
        PhySG~\cite{zhanf2021physg} & 0.70 \% & 0.49 \% \\
        NDRMC~\cite{hasselgren2022nvdiffrecmc} & 0.64 \% & 0.44 \% \\
        \rowcolor[rgb]{0.93,1.0,0.87} Ours & \textbf{0.15 \%} & \textbf{0.15 \%}  \\
    \end{tabular}
    \label{tab:geometry-results}
  }
  ~
  \subfloat[][]{
  \setlength{\tabcolsep}{3pt}
  \begin{tabular}{l|cr}
    & RMS1 & RMS2 \\ \hline
    Initial & 1.10 \% & 1.03 \%  \\
    w/o SfS & 0.16 \% & 0.15 \% \\
    w/o NeRF & 0.21 \% & 0.20 \%  \\
    \rowcolor[rgb]{0.93,1.0,0.87} Ours & \textbf{0.15} \% & \textbf{0.15} \%
    \label{tab:geometry-results-ablation}
  \end{tabular}
  }
  \caption{Geometry reconstruction accuracy (RMSE) on nLMVS-Synth dataset \cite{yamashita2023nlmvs}. RMS1 is the root-mean-square error (RMSE) from the recovered surface to the nearest ground truth surface and RMS2 is RMSE from the ground truth surface to the nearest recovered surface. }
\end{table}

\begin{table*}[t]
  \centering
  \small
  \subfloat[][RMSE from recovered to ground truth (RMS1)]{
      \setlength{\tabcolsep}{3.7pt}
    \begin{tabular}{l|cccccc}
        Object & Buildings-Metal-Horse & Court-Metal-Shell & Entrance-White-Horse & Labo-Blue-Shell & Labo-Metal-Bunny & Mean \\ \hline
        IDR~\cite{yariv2020idr} & \textbf{0.37 \%} & 0.49 \% & \textbf{0.38} \% & 0.40 \% & 0.95 \% & 0.52 \% \\
        NeuS~\cite{wang2021neus} & 3.14 \% & 2.70 \% & 0.60 \% & 1.51 \% & 1.00 \% & 1.79 \% \\
        PhySG~\cite{zhanf2021physg} & 0.70 \% & 1.01 \% & 0.69 \% & 0.57 \% & 0.90 \% & 0.78 \% \\
        NDRMC~\cite{hasselgren2022nvdiffrecmc} & 1.06 \% & 1.61 \% & 0.74 \% & 0.92 \% & 2.32 \% & 1.33 \% \\
        NeRO~\cite{liu2023nero} & 2.37 \% & 1.51 \% & 0.62 \% & \textbf{0.23 \%} & 1.51 \% & 1.25 \% \\
        \rowcolor[rgb]{0.93,1.0,0.87} Ours & 0.42 \% & \textbf{0.36 \%} & 0.41 \% & 0.27 \% & \textbf{0.33} \% & \textbf{0.36} \% \\
    \end{tabular}
  }
  \\
  \subfloat[][RMSE from ground truth to recovered (RMS2)]{
  \setlength{\tabcolsep}{3.7pt}
    \begin{tabular}{l|cccccc}
        Object & Buildings-Metal-Horse & Court-Metal-Shell & Entrance-White-Horse & Labo-Blue-Shell & Labo-Metal-Bunny & Mean \\ \hline
        IDR~\cite{yariv2020idr} & \textbf{0.36} \% & 0.49 \% & 0.41 \% & 0.34 \% & 0.81 \% & 0.48 \% \\
        NeuS~\cite{wang2021neus} & 1.75 \% & 1.99 \% & 0.51 \% & 0.88 \% & 0.68 \% & 1.16 \% \\
        PhySG~\cite{zhanf2021physg} & 0.66 \% & 0.68 \% & 0.68 \% & 0.48 \% & 0.80 \% & 0.66 \% \\
        NDRMC~\cite{hasselgren2022nvdiffrecmc} & 0.76 \% & 1.11 \% & 0.46 \% & 0.68 \% & 0.99 \% & 0.80 \% \\
        NeRO~\cite{liu2023nero} & 2.61 \% & 1.68 \% & 0.59 \% & \textbf{0.24 \%} & 3.86 \% & 1.80 \% \\
        \rowcolor[rgb]{0.93,1.0,0.87} Ours & 0.39 \% & \textbf{0.36} \% & \textbf{0.38 \%} & 0.34 \% & \textbf{0.37 \%} & \textbf{0.37 \%} \\
    \end{tabular}
  }
  \caption{Geometry reconstruction accuracy (RMSE) on nLMVS-Real dataset~\cite{yamashita2023nlmvs}. The results demonstrate the robustness of our method. }
  \label{tab:geometry-results-real}
\end{table*}

\section{Experimental Results}
 
We focus our experiments on answering the main questions below. Can the RM network recover accurate reflectance maps? Does the joint estimation framework improve geometry reconstruction accuracy? How does our method work compare with inverse rendering~\cite{zhanf2021physg, hasselgren2022nvdiffrecmc, liu2023nero} and neural image synthesis~\cite{yariv2020idr, wang2021neus} methods? How does our method generalize to real data? How does each component of our model contribute to the estimation accuracy? To answer these questions, we validate the following points.
\begin{itemize}
    \item Estimation results of RM-Net are qualitatively and quantitatively more accurate compared to baseline methods.
    \item Our joint estimation framework improves the accuracy of the recovered geometry.
    \item The recovered 3D shapes are more accurate than those of existing methods on both synthetic and real data.
\end{itemize}

\paragraph{Training Data} We used the nLMVS-Synth dataset~\cite{yamashita2023nlmvs} for training of the proposed networks. The training dataset consists of 26850 images of 2685 combinations of shapes, materials, and illuminations. We applied random rotation and zoom to the training images as data augmentation.

\subsection{Accuracy of RM Network}

We first evaluated the accuracy of the RM Network using images and (ground truth) surface normal maps from the test set of the nLMVS-Synth dataset~\cite{yamashita2023nlmvs} as inputs. To clarify the effectiveness of the proposed image-space feature extraction and pixel-wise confidence estimation, we compared our method with its three variants, ``w/o Confidence'', ``w/o Image Feature'', and ``Naive''. ``w/o Confidence'' does not recover the pixel-wise confidence scores and uses a constant weight during the weighted mapping. ``w/o Image Feature'' does not extract the image features and input pixel values are used as alternatives. ``Naive'' does not use either the image feature or the pixel-wise score similar to Georgoulis \etal~\cite{georgoulis18deeprani}. \Cref{fig:rm-results} and \Cref{tab:rm-results} show qualitative and quantitative results. In \cref{fig:rm-results}, as a reference, we also show a sparse reflectance map obtained by simply mapping pixels in the input image using the normal map. The results clearly show the effectiveness of the DeepShaRM.

\subsection{Joint Shape and Reflectance Map Recovery}
\label{sec:exp-joint-est}

We evaluated the accuracy of the overall joint shape and reflectance map estimation on the test set of nLMVS-Synth dataset. We used 10 view images as inputs for each object. We first compared our method with its two variants, ours without the SfS loss (``w/o SfS loss'') and ours without the Synthesis loss (``w/o NeRF''), to analyze the effectiveness of each of the loss functions. We quantitatively evaluated the accuracy of the recovered geometry by computing the root-mean-square (RMS) distance between the recovered and ground truth 3D shapes. We computed this error using only the surfaces visible from the input views. \Cref{fig:ablation-results} and \cref{tab:geometry-results-ablation} show qualitative and quantitative results. In \cref{tab:geometry-results-ablation}, we also show the accuracy of the initial geometry estimate (\ie, visual hull). The results show that the proposed SfS loss is essential to obtain plausible results especially when the surface is highly specular, while the NeRF improves the reconstruction of surface details. We also compared our method with recent inverse rendering~\cite{zhanf2021physg, hasselgren2022nvdiffrecmc, liu2023nero} and neural image synthesis~\cite{yariv2020idr, wang2021neus} methods. \Cref{fig:opening} and \cref{tab:geometry-results} show qualitative and quantitative results. Our method achieves the state-of-the-art accuracy on the nLMVS-Synth dataset~\cite{yamashita2023nlmvs}. Note that, while NeRO~\cite{liu2023nero} can jointly recover silhouettes of the object and surface geometry, we found that they often completely fail on the 10 view inputs. For this reason, we couldn't quantitatively compare our method with NeRO~\cite{liu2023nero}. Note also that, as shown in \cref{fig:opening}, even when NeRO~\cite{liu2023nero} successfully estimates silhouettes of the object, our geometry estimation result is more accurate than that by NeRO~\cite{liu2023nero}.

\subsection{Evaluation on Real Data}

We also evaluated DeepShaRM on real-world images from the nLMVS-Real Dataset~\cite{yamashita2023nlmvs}. \Cref{fig:geometry-results-real} and \cref{tab:geometry-results-real} show qualitative and quantitative results of our method and those of existing methods. Our method successfully recovers accurate geometry on these real-world images. This clearly demonstrates the robustness our method. As we estimate camera-view reflectance maps separately for each view, our method can deal with real-world outdoor scenes where the surrounding illumination (\eg, sunlight) can dynamically change.

\Cref{fig:rm-results-real} shows recovered reflectance maps from the real-world images in the nLMVS-Real dataset. Although the ground truth reflectance maps are not available, the results are qualitatively consistent with corresponding reflectance maps of mirror objects created from ground truth illumination maps. 

\subsubsection{Estimation from Very Sparse Inputs}

We also tested our method on a very sparse setup, \ie, estimation from only five view real-world images (one image and its four neighboring images). \Cref{fig:geometry-results-sparse} shows qualitative results. Our method can work even on these very sparse in-the-wild inputs. This further demonstrates the effectiveness and robustness of our method.

\section{Conclusion}
We introduced DeepShaRM, a novel deep multi-view shape and reflectance map estimation method for textureless, non-Lambertian objects under unknown natural illumination. By bypassing the ill-posed decomposition of surface appearance into reflectance and illumination and instead estimating camera-view reflectance maps with strong learned priors, accurate 3D shapes are successfully recovered from posed multi-view images without making any restricting assumptions (\eg, known illumination). Experimental results including evaluation on real-world images demonstrate state-of-the-art accuracy and robustness to changing illumination and global light transport effects both of which are inevitable in the wild. We believe our method offers a useful practical means for passive 3D reconstruction in the wild.

\vspace{-6pt}
\paragraph*{Acknowledgement}
This work was in part supported by
JSPS 
20H05951, 
21H04893, 
23H03420, 
JST
JPMJCR20G7, 
JPMJSP2110, 
and RIKEN GRP.

{
    \small
    \bibliographystyle{ieeenat_fullname}
    \bibliography{3dv24_kyamashita}
}

\clearpage
\setcounter{page}{1}
\maketitlesupplementary

\appendix

\section{Implementation Details}

\subsection{RM Network}
We use a UNet architecture described in \cref{tab:rm_unet_arch} to implement the sub-networks of the RM network. For the ``Feature Net'', the ``Initial Confidence Net'', and the ``Confidence Net'', we use the UNet architecture as is. We set the number of output channels of the ``Feature Net'' to 32. We apply the differentiable mapping described in \cref{eq:mapping} to each output channel of the ``Feature Net''. As the outputs of the ``Initial Confidence Net'' and the ``Confidence Net'' are pixel-wise confidence scores, we set the number of output channels of these networks to 1. We apply sigmoid activation to the outputs of these networks so that they are in the interval $[0, 1]$. For the ``RM Decoder'', we stack two UNets and use the output of the first UNet as an additional input of the second one. We apply the exponential function to the output of the ``RM Decoder'' so that pixel values in the output reflectance map become positive.

As mentioned in the main paper, we alternate between estimating reflectance maps and refining the pixel-wise confidence scores. We iterate the alternating estimation three times.

\begin{table}
  \centering
  \small
  \begin{tabular}{l|cccl}
     Layer & k & s & chns & input \\ \hline
     $\mathrm{CBR2D_0}$ & 3 & 1 & in\_chns/64 & Original Input \\
     $\mathrm{AP_0}$ & - & 2 & 64/64 & $\mathrm{CBR2D_0}$ \\
     $\mathrm{CBR2D_1}$ & 3 & 1 & 64/128 & $\mathrm{AP_0}$ \\
     $\mathrm{CBR2D_2}$ & 3 & 1 & 128/128 & $\mathrm{CBR2D_1}$ \\
     $\mathrm{AP_1}$ & - & 2 & 128/128 & $\mathrm{CBR2D_2}$ \\
     $\mathrm{CBR2D_3}$ & 3 & 1 & 128/256 & $\mathrm{AP_1}$ \\
     $\mathrm{CBR2D_4}$ & 3 & 1 & 256/256 & $\mathrm{CBR2D_3}$ \\
     $\mathrm{AP_2}$ & - & 2 & 256/256 & $\mathrm{CBR2D_4}$ \\
     $\mathrm{CBR2D_5}$ & 3 & 1 & 256/512 & $\mathrm{AP_2}$ \\
     $\mathrm{CBR2D_6}$ & 3 & 1 & 512/512 & $\mathrm{CBR2D_5}$ \\
     $\mathrm{CBR2D_7}$ & 3 & 1 & 512/256 & $\mathrm{CBR2D_6}$ \\
     $\mathrm{UP_0}$ & - & - & 256/256 & $\mathrm{CBR2D_7}$ \\
     $\mathrm{CBR2D_8}$ & 3 & 1 & (256+256)/128 & $\mathrm{CBR2D_4}$, $\mathrm{UP_0}$ \\
     $\mathrm{UP_1}$ & - & - & 128/128 & $\mathrm{CBR2D_8}$ \\
     $\mathrm{CBR2D_9}$ & 3 & 1 & (128+128)/64 & $\mathrm{CBR2D_2}$, $\mathrm{UP_1}$ \\
     $\mathrm{UP_2}$ & - & - & 64/64 & $\mathrm{CBR2D_9}$ \\
     $\mathrm{Conv2D}$ & 3 & 1 & (64+64)/out\_chns & $\mathrm{CBR2D_0}$, $\mathrm{UP_2}$ \\
  \end{tabular}
\caption{The UNet architecture we use to implement sub-networks in the RM Network and the SfS Network (the same as \cite{yamashita2023nlmvs}). ``CBR2D'' is a sequence consisting of 2D convolution (``Conv2D''), batch normalization, and ReLU activation. ``AP'' and ``UP'' are average pooling and bilinear upsampling, respectively, that double or halve the size of an input feature map. ``k'', ``s'', and ``chns'' denote the kernel size, the stride, and the numbers of input and output channels, respectively. ``in\_chns'' is the number input channels and ``out\_chns'' is the number of the output channels.}
  \label{tab:rm_unet_arch}
\end{table}

\begin{figure*}[t]
  \centering
  \includegraphics[keepaspectratio, width=\linewidth]{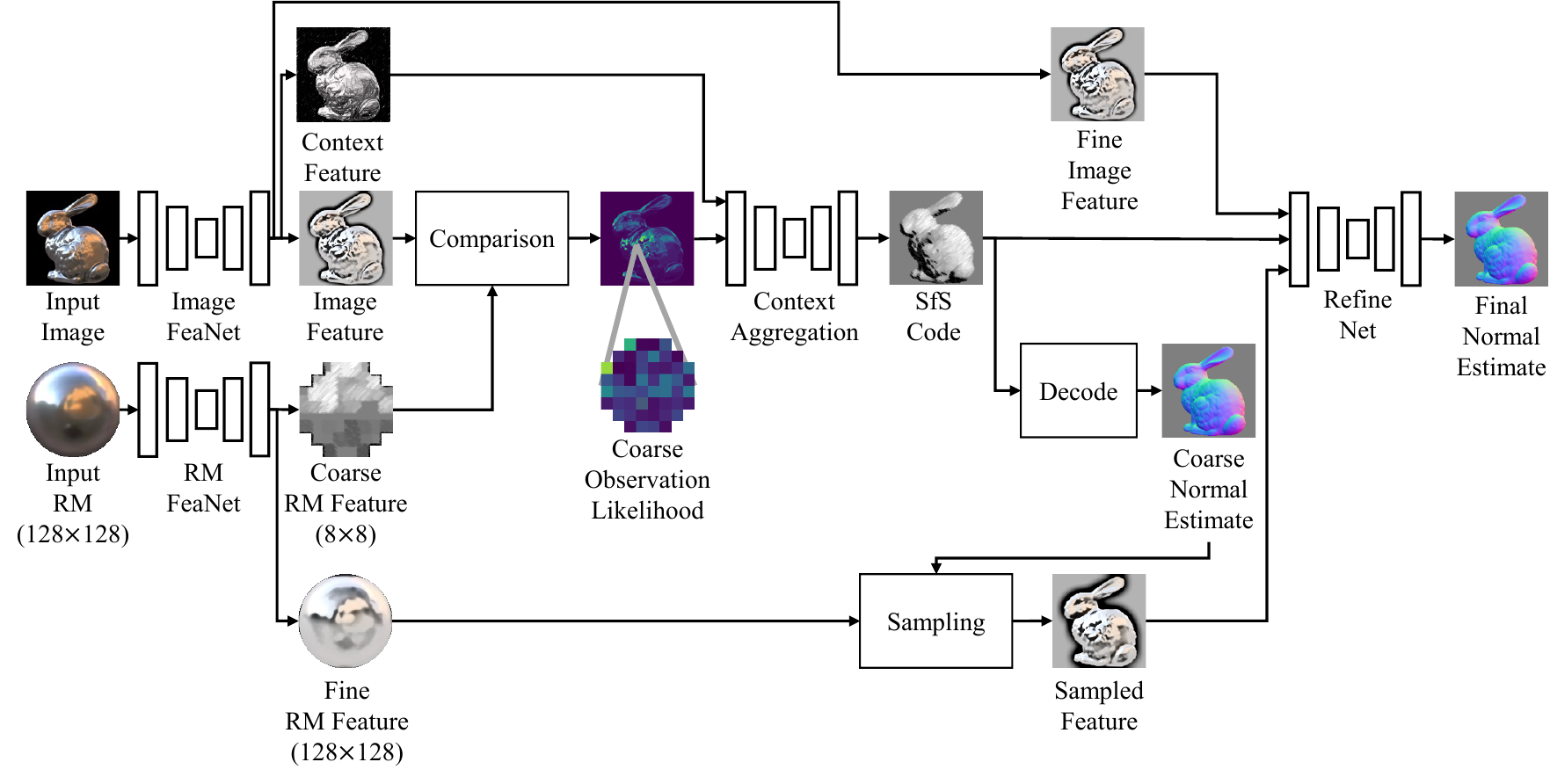}
  \caption{The architecture of the SfS network. To reduce the computational burden, we leverage the input reflectance map (RM) in a coarse-to-fine manner. We first compute a pixel-wise coarse observation likelihood for a small number of surface normal candidates using a low-resolution feature reflectance map extracted from the input reflectance map. Once the surface normals are estimated using the coarse observation likelihood, we refine them by leveraging features that we sample from a fine (high resolution) feature reflectance map using the coarse surface normal estimate.
  }
  \label{fig:sfs-net-details}
\end{figure*}

We use synthetic images, synthetic (ground truth) normal maps, and corresponding ground truth reflectance maps from the nLMVS-Synth dataset~\cite{yamashita2023nlmvs} for training RM Network. We train the network by minimizing the errors between the reflectance maps recovered from the inputs and the corresponding ground truth. Specifically, we evaluate the error between an estimated reflectance map $\hat{R}(\mathbf{n})$ and a corresponding ground truth $R(\mathbf{n})$ using log-space L1 loss
\begin{equation}
    L_\mathrm{L1} =  \sum_{\mathbf{n} \in N} \left\|\log \hat{R}(\mathbf{n}) - \log R(\mathbf{n})\right\|_1 \,,
\end{equation}
log-space gradient loss
\begin{equation}
    L_\mathrm{G} = \sum_{\mathbf{n} \in N} \left\|g\left(\log \hat{R}(\mathbf{n})\right) - g\left(\log R(\mathbf{n})\right)\right\|_1 \,,
\end{equation}
and perceptual loss
\begin{equation}
    L_\mathrm{P} = \sum_{\mathbf{n} \in N} \left\|f\left(\hat{R}(\mathbf{n})\right) - f\left(R(\mathbf{n})\right)\right\|_2 \,,
\end{equation}
where $N$ is a set of surface normals that correspond to pixels in the reflectance maps, $g(\cdot)$ is an operation that computes gradients of the input feature map with respect to horizontal and vertical directions (\ie, the number of channels is doubled), and $f(\cdot)$ is a feature extraction layer of a pretrained VGG16 model. Note that we can apply the VGG16 feature extraction layer to the reflectance maps as we represent them as 2D images using the angular fisheye projection. We also compute an image reconstruction loss 
\begin{equation}
    L_\mathrm{I} = \sum_\mathbf{m} \alpha_\mathbf{m} \left\| \log\hat{R}(\mathbf{n_m}) - \log I_\mathbf{m} \right\|_1 \,,
\end{equation}
where $I_\mathbf{m}$ and $\mathbf{n_m}$ are the input pixel value and the input surface normal of a pixel $\mathbf{m}$. We compute $\alpha_\mathbf{m}$ 
\begin{equation}
    \alpha_\mathbf{m} = \exp\left(-10 \left\| \log R(\mathbf{n_m}) - \log I_\mathbf{m} \right\|_1\right)\,,
\end{equation}
so that we can impose this loss only on pixels consistent with the ground truth reflectance map. We use the weighted sum of these loss functions 
\begin{equation}
    L_\mathrm{RM} = w_\mathrm{L1} L_\mathrm{L1} + w_\mathrm{G} L_\mathrm{G} + w_\mathrm{P} L_\mathrm{P} + w_\mathrm{I} L_\mathrm{I} \,,
\end{equation}
as the training loss for the RM network. In practice, we set the weights $w_\mathrm{L1}$, $w_\mathrm{G}$, and $w_\mathrm{P}$, and $w_\mathrm{I}$ to 1, 0.1, 0.001, and 1, respectively. As the RM network outputs a reflectance map for each iteration of the alternating estimation, we compute this loss for the output of each iteration.

\subsection{SfS Network}

\Cref{fig:sfs-net-details} depicts details of the architecture of the SfS network. As mentioned in the main paper, inspired by Yamashita \etal~\cite{yamashita2023nlmvs}, we compute the pixel-wise observation likelihood for each surface normal candidate by comparing the input image and the input reflectance map. Naive computation for all surface normal candidates is, however, computationally costly. Although Yamashita \etal~\cite{yamashita2023nlmvs} has reduced this cost by introducing a coarse-to-fine sampling method of the surface normal candidates, we found that they still require a large memory footprint. As we use the SfS network for each iteration in the joint estimation, the computational burden is critical. For this, we further simplify the network architecture by introducing another coarse-to-fine estimation method. Given the input image and the input reflectance map, we first compute the observation likelihood for a small number of surface normal candidates using a low-resolution ($8\times8$) coarse feature reflectance map. Specifically, we compute the observation likelihood $p_\mathbf{m}(\mathbf{n})$ using a feature map $I'(\mathbf{m})$ extracted from the input image and the coarse reflectance map feature $R'_c(\mathbf{n})$ extracted from the input reflectance map 
\begin{equation}
    p_\mathbf{m}(\mathbf{n}) = \frac{\exp\left(-\left\|I'(\mathbf{m}) - R_c'(\mathbf{n})\right\|_2\right)}{\sum_{\mathbf{n'} \in N_c} \exp\left(-\left\|I'(\mathbf{m}) - R_c'(\mathbf{n'})\right\|_2\right)} \,,
\end{equation}
where $N_c$ is a set of 64 surface normal candidates that correspond to pixels in the coarse reflectance map feature. We compute this likelihood for the 64 surface normal candidates and treat it as a 2D feature map that has 64 channels. We feed this likelihood along with another feature map extracted from the input image to a 2D convolutional sub-network (Context Aggregation in the figure) that outputs a coarse surface normal estimate (SfS code) by leveraging contextual information. We represent the coarse estimate as pixel-wise mixtures of von Mises-Fisher (vMF) distributions so that it can encode ambiguity of the coarse estimation. The network outputs a mixing weight $\pi_k(\mathbf{m})$, a mean direction $\mathbf{\mu_k}(\mathbf{m})$, and a concentration parameter $\kappa_k(\mathbf{m})$ for each pixel $\mathbf{m}$ and each vMF distribution indexed by $k$. The estimate can also be decoded into a single surface normal estimate for each pixel by computing the mean direction
\begin{equation}
    \hat{\mathbf{n_0}}(\mathbf{m}) = \frac{\sum_k \pi_k(\mathbf{m}) \mathbf{\mu_k}(\mathbf{m})}{\left\|\sum_k \pi_k(\mathbf{m}) \mathbf{\mu_k}(\mathbf{m})\right\|_2} \,.
\end{equation}

Once the coarse estimate is obtained, we can further refine it using another fine (high resolution) reflectance map feature extracted from the input reflectance map. We sample features $I'_f(\mathbf{m})$ from the fine reflectance map feature $R'_f(\mathbf{n})$ using the coarse surface normal estimate $\hat{\mathbf{n_0}}(\mathbf{m})$ 
\begin{equation}
    I'_f(\mathbf{m}) = R'_f\left(\hat{\mathbf{n_0}}(\mathbf{m})\right) \,,
\end{equation}
and use them as inputs to a refinement network that outputs the final estimates as pixel-wise surface normals $\hat{\mathbf{n}}(\mathbf{m})$.

Similar to the RM network, we train the SfS network in a supervised manner. We use synthetic images and synthetic reflectance maps as inputs and compare the outputs with ground truth surface normal maps. We evaluate consistency between the surface normal estimates (\ie, $\hat{\mathbf{n_0}}(\mathbf{m})$ and $\hat{\mathbf{n}}(\mathbf{m})$) and the ground truth by computing mean absolute errors as loss functions. We also impose consistency between the vMF estimates and the ground truth by computing a negative log likelihood loss
\begin{equation}
    L_\mathbf{NLL} = -\log\left(\sum_k \pi_k(\mathbf{m}) f_p(\mathbf{n_t}(\mathbf{m}); \mathbf{\mu_k}(\mathbf{m}), \kappa_k(\mathbf{m}))\right) \,,
\end{equation}
where $\mathbf{n_t}(\mathbf{m})$ is the ground truth surface normal and $f_p(\mathbf{x}; \mathbf{\mu}, \kappa)$ is the probability density function of the vMF distribution. We use the sum of these loss functions as the training loss.

\begin{table}
  \centering
  \small
  \begin{tabular}{l|cccl}
     Layer & k & s & chns & input \\ \hline
     $\mathrm{CBR2D_0}$ & 3 & 1 & 3/64 & Input RM \\
     $\mathrm{AP_0}$ & - & 2 & 64/64 & $\mathrm{CBR2D_0}$ \\
     $\mathrm{CBR2D_1}$ & 3 & 1 & 64/128 & $\mathrm{AP_0}$ \\
     $\mathrm{CBR2D_2}$ & 3 & 1 & 128/128 & $\mathrm{CBR2D_1}$ \\
     $\mathrm{AP_1}$ & - & 2 & 128/128 & $\mathrm{CBR2D_2}$ \\
     $\mathrm{CBR2D_3}$ & 3 & 1 & 128/256 & $\mathrm{AP_1}$ \\
     $\mathrm{CBR2D_4}$ & 3 & 1 & 256/256 & $\mathrm{CBR2D_3}$ \\
     $\mathrm{AP_2}$ & - & 2 & 256/256 & $\mathrm{CBR2D_4}$ \\
     $\mathrm{CBR2D_5}$ & 3 & 1 & 256/512 & $\mathrm{AP_2}$ \\
     $\mathrm{CBR2D_6}$ & 3 & 1 & 512/512 & $\mathrm{CBR2D_5}$ \\
     $\mathrm{CBR2D_7}$ & 3 & 1 & 512/256 & $\mathrm{CBR2D_6}$ \\
     $\mathrm{UP_0}$ & - & - & 256/256 & $\mathrm{CBR2D_7}$ \\
     $\mathrm{CBR2D_8}$ & 3 & 1 & (256+256)/128 & $\mathrm{CBR2D_4}$, $\mathrm{UP_0}$ \\
     $\mathrm{UP_1}$ & - & - & 128/128 & $\mathrm{CBR2D_8}$ \\
     $\mathrm{CBR2D_9}$ & 3 & 1 & (128+128)/64 & $\mathrm{CBR2D_2}$, $\mathrm{UP_1}$ \\
     $\mathrm{UP_2}$ & - & - & 64/64 & $\mathrm{CBR2D_9}$ \\
     $\mathrm{Conv2D_0}$ & 3 & 1 & (64+64)/32 & $\mathrm{CBR2D_0}$, $\mathrm{UP_2}$ \\
     $\mathrm{AP_3}$ & - & 2 & 512/512 & $\mathrm{CBR2D_6}$ \\
     $\mathrm{CBR2D_{10}}$ & 3 & 1 & 512/512 & $\mathrm{AP_3}$ \\
     $\mathrm{Conv2D_1}$ & 3 & 1 & 512/32 & $\mathrm{CBR2D_{10}}$ \\
  \end{tabular}
\caption{The architecture of ``RM FeaNet'' in the SfS network. We use the output of $\mathrm{Conv2D_0}$ as the fine RM feature and the output of $\mathrm{Conv2D_1}$ as the coarse RM feature.}
  \label{tab:rm_feanet_arch}
\end{table}

\begin{table}
  \centering
  \small
  \begin{tabular}{l|cccl}
     Layer & k & s & chns & input \\ \hline
     $\mathrm{CBR2D_0}$ & 1 & 1 & in\_chns/256 & Input Feature \\
     $\mathrm{CBR2D_1}$ & 1 & 1 & 256/256 & $\mathrm{CBR2D_0}$ \\
     $\mathrm{CBR2D_2}$ & 1 & 1 & 256/256 & $\mathrm{CBR2D_1}$ \\
     $\mathrm{CBR2D_3}$ & 1 & 1 & 256/256 & $\mathrm{CBR2D_2}$ \\
     $\mathrm{Conv2D}$ & 3 & 1 & 256/2 & $\mathrm{CBR2D_3}$ \\
  \end{tabular}
\caption{We implement the ``Refine Net'' in the SfS network by stacking the UNet described in \cref{tab:rm_unet_arch} and a shallow convolutional neural network described in this table. In addition to the outputs of the UNet, we use the original inputs (inputs of the UNet) as additinal inputs to this subsequent network.}
  \label{tab:refine_net_arch}
\end{table}

\subsubsection{Details of Network Architecture}
Similar to the RM network, we use the UNet architecture described in \cref{tab:rm_unet_arch} to implement the ``Image FeaNet'' and the ``Context Aggregation''. We set the number of channels of the image feature, the fine image feature, and the context feature to 32. Thus the number of output channels of ``Image FeaNet'' is 96. As described in \cref{tab:rm_feanet_arch}, we use a slightly different architecture for the ``RM FeaNet''. \Cref{tab:refine_net_arch} shows the architecture of ``Refine Net''. The ``Refine Net'' outputs two feature maps $p(\mathbf{m})$ and $q(\mathbf{m})$ that we convert to the output surface normal $\hat{\mathbf{n}}(\mathbf{m}) = (\hat{n_x}(\mathbf{m}),\hat{n_y}(\mathbf{m}),\hat{n_z}(\mathbf{m}))^\mathrm{T}$ 
\begin{equation}
    \hat{n_x}(\mathbf{m}) = \frac{p(\mathbf{m})}{l} \,,
\end{equation}
\begin{equation}
    \hat{n_y}(\mathbf{m}) = \frac{q(\mathbf{m})}{l} \,,
\end{equation}
\begin{equation}
    \hat{n_z}(\mathbf{m}) = \frac{1}{l} \,,
\end{equation}
where
\begin{equation}
    l = \sqrt{p(\mathbf{m})^2 + q(\mathbf{m})^2+1} \,.
\end{equation}
This ensures that the output surface normal always satisfies $\|\hat{\mathbf{n}}(\mathbf{m})\|_2 = 1$ and $\hat{n_z}(\mathbf{m}) > 0$.

\begin{figure*}[t]
  \centering
  \includegraphics[keepaspectratio, width=\linewidth]{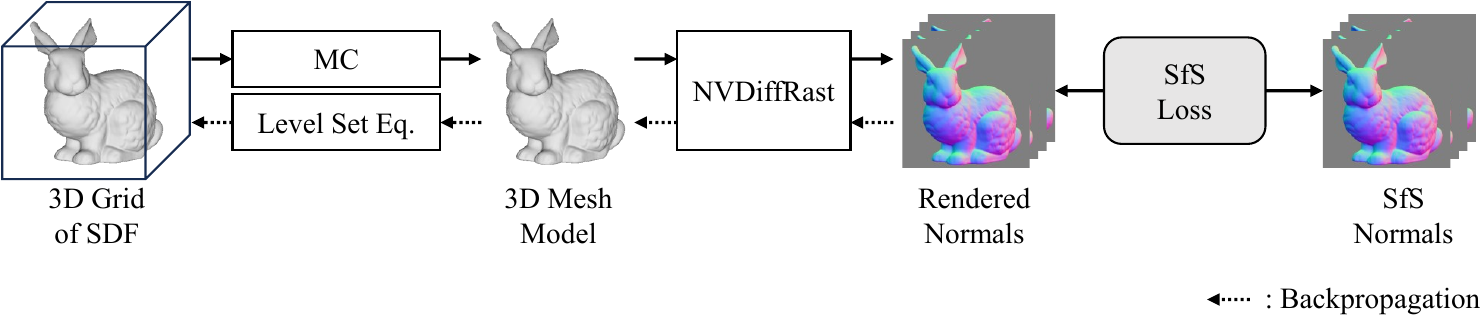}
  \caption{Inspired by Neural Implicit Evolution~\cite{mehta2022nie}, we update the geometry estimate represented as a 3D grid of a signed distance function using a level set method and NVDiffRast~\cite{laine20nvdiffrast}, a differentiable renderer for 3D mesh models.
  }
  \label{fig:sdf-optimization}
\end{figure*}

\subsection{SDF Optimization}
\Cref{fig:sdf-optimization} depicts how we update the 3D grid of a signed distance function (SDF) using surface normals estimated by the SfS Network. As described in the main paper, we first extract a 3D mesh model from the SDF grid using Marching Cubes~\cite{lorensen87marchingcubes} and compute gradients of the SfS loss with respect to vertex positions of the mesh model using NVDiffRast~\cite{laine20nvdiffrast}, a differentiable renderer for 3D mesh models. We then compute gradients with respect to parameters of the SDF using the level set equation and use them to update the SDF parameters. That is, given the gradients $\frac{\partial L}{\partial \mathbf{v_j}}$ with respect to vertex positions $\mathbf{v_j}$, we compute the gradients with respect to the SDF values $f_\theta(\mathbf{v_j})$ 
\begin{equation}
    \frac{\partial L}{\partial f_\theta(\mathbf{v_j})} = -\nabla f_\theta(\mathbf{v_j}) \cdot \frac{\partial L}{\partial \mathbf{v_j}} \,,
\end{equation}
and then compute the gradients with respect to parameters of the SDF $\theta$ (\ie, values in the 3D grid) 
\begin{equation}
    \frac{\partial L}{\partial \theta} = \sum_j \frac{\partial f_\theta(\mathbf{v_j})}{\partial \theta}\frac{\partial L}{\partial f_\theta(\mathbf{v_j})} \,,
\end{equation}
where $j$ is index of the vertex. Note that we compute the SDF value $f_\theta(\mathbf{v_k})$ from the parameters $\theta$ using the cubic B-spline interpolation and compute the gradient $\frac{\partial f_\theta(\mathbf{v})}{\partial\theta}$ by automatic differentiation. We use the obtained gradient $\frac{\partial L}{\partial \theta}$ and the Adam optimizer to update the SDF estimate. 

As mentioned in the main text, optionally, we also synthesize images from the SDF estimate and a neural radiance field (NeRF)~\cite{mildenhall2020nerf} and jointly optimize their parameters by minimizing errors between the synthesized images and the inputs. For this, in addition to the rendered normal map $\bar{\mathbf{n}}(\mathbf{m})$, we also render 2D maps of surface positions (\ie, intersections of camera rays and the surface) $\bar{\mathbf{x}}(\mathbf{m})$ and viewing directions $\bar{\mathbf{v}}(\mathbf{m})$ for each view using NVDiffRast. Given these and a NeRF $c(\mathbf{x},\mathbf{v})$, an MLP that takes in a 3D position and a viewing direction and outputs a surface radiance, we compute the synthesized image $\bar{I}(\mathbf{m})$ 
\begin{equation}
    \bar{I}(\mathbf{m}) = c(\mathbf{x}(\mathbf{m}),\mathbf{v}(\mathbf{m})) \,.
\end{equation}
We evaluate the consistency between the synthesized and the input images using the log-scale L1 loss. We also render silhouette images using NVDiffRast and evaluate its consistency with the input silhouette images using the smooth L1 loss.

\begin{figure*}[t]
  \centering
  \includegraphics[keepaspectratio, width=\linewidth]{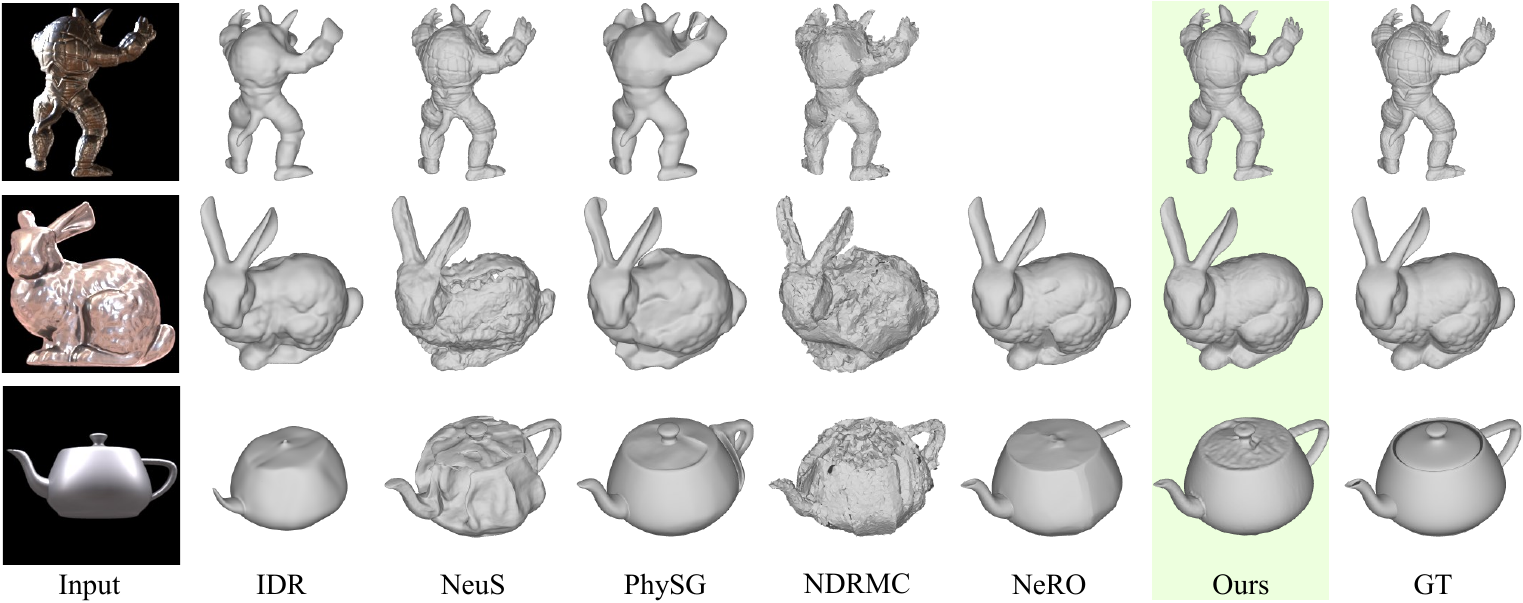}
  \caption{Geometry estimation results on nLMVS-Synth Dataset~\cite{yamashita2023nlmvs}. NeRO~\cite{liu2023nero} completely failed to recover the 3D shape of the object in the first row.
  }
  \label{fig:geometry-results-synth-supp}
\end{figure*}

\begin{figure*}[t]
  \centering
  \includegraphics[keepaspectratio, width=\linewidth]{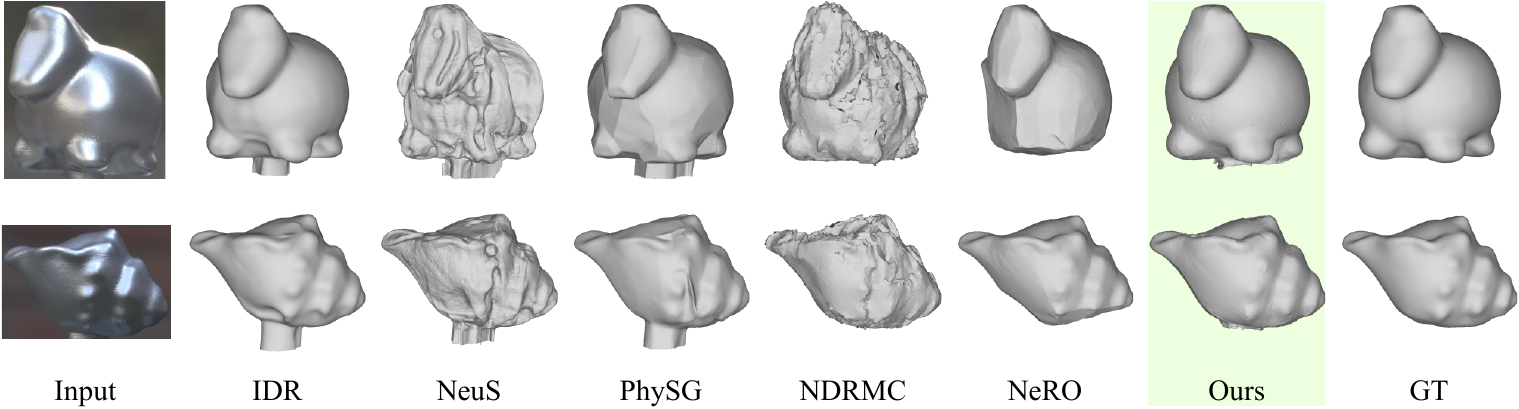}
  \caption{Additional geometry estimation results on the nLMVS-Real Dataset~\cite{yamashita2023nlmvs}.
  }
  \label{fig:geometry-results-real-supp}
\end{figure*}

\begin{figure}[t]
  \centering
  \includegraphics[keepaspectratio, width=\linewidth]{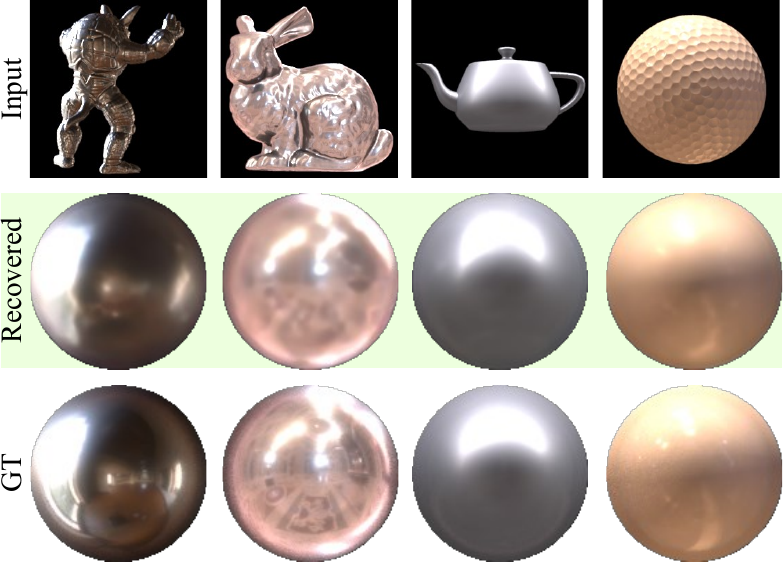}
  \caption{Reflectance maps recovered by the joint estimation of geometry and reflectance maps on the nLMVS-Synth dataset~\cite{yamashita2023nlmvs}. The results are qualitatively consistent with the ground truth.
  }
  \label{fig:rm-results-synth}
\end{figure}

\section{Additional Qualitative Results}
\Cref{fig:geometry-results-synth-supp} and \cref{fig:geometry-results-real-supp} show qualitative geometry estimation results on the nLMVS-Synth and the nLMVS-Real datasets~\cite{yamashita2023nlmvs}, respectively. Our method works well on objects with different shapes and materials, and also under different natural illumination environments.

\Cref{fig:rm-results-synth} shows reflectance maps recovered by the joint estimation of geometry and reflectance maps on the nLMVS-Synth dataset~\cite{yamashita2023nlmvs}. The results are qualitatively consistent with the ground truth.

\end{document}